\documentclass[letterpaper, 10 pt, conference]{ieeeconf}  

\IEEEoverridecommandlockouts                              

\usepackage{multicol}
\usepackage{booktabs}
\usepackage[table]{xcolor}

\usepackage{graphicx}
\usepackage{duckuments}
\usepackage{amsmath}

\usepackage{xcolor}
\usepackage{todonotes}

\usepackage{pifont}
\usepackage{bbding}

\usepackage{subcaption}

\usepackage[acronym]{glossaries}
\glsdisablehyper
\newacronym{vpr}{VPR}{Visual Place Recognition}
\newacronym{dl}{DL}{Deep Learning}
\newacronym{slam}{SLAM}{Simultaneous Localisation And Mapping}
\newacronym{sfm}{SfM}{Structure from Motion}
\newacronym{emd}{EMD}{Earth-Mover Distance}
\newacronym{cnn}{CNN}{Convolutional Neural Network}
\newacronym[firstplural=Visual Distributions of Neuron Activations (VDNAs)]{vdna}{VDNA}{Visual Distribution of Neuron Activations}
\newacronym{fid}{FID}{Fr\'echet Inception Distance}
\newacronym{kid}{KID}{Kernel Inception Distance}
\newacronym{fd}{FD}{Fr\'echet Distance}
\newacronym{pdf}{PDF}{probability distribution function}
\newacronym{cdf}{CDF}{cumulative distribution function}
\newacronym{vit}{ViT}{Vision Transformer}
\newacronym{bow}{BoW}{bag-of-words}

\def\etal{\emph{et al}.}

\usepackage[labelformat=simple]{subcaption}
\DeclareCaptionLabelSeparator{periodspace}{.\quad}
\usepackage{lipsum}

\usepackage{crossreftools}

\makeatletter
\let\NAT@parse\undefined
\makeatother

\usepackage[hidelinks]{hyperref}
\newcommand\rurl[1]{%
\texttt{\href{http://#1}{\nolinkurl{#1}}}%
}
\usepackage[capitalise]{cleveref}

\title{\LARGE \bf
\textit{VDNA-PR}: Using General Dataset Representations \\for Robust Sequential Visual Place Recognition
}
\author{Benjamin Ramtoula$^{*}$, Daniele De Martini$^+$, Matthew Gadd$^+$, and Paul Newman
\\
Mobile Robotics Group (MRG), University of Oxford\\\texttt{\{benjamin,daniele,mattgadd,pnewman\}@robots.ox.ac.uk}
\thanks{$^{*}$ Corresponding author.}
\thanks{$^+$ Equal contribution.}
\thanks{
This work was supported by EPSRC Programme Grant ``From Sensing to Collaboration'' (EP/V000748/1), the EPSRC Centre for Doctoral Training in Autonomous Intelligent Machines and Systems [EP/S024050/1], and Oxa.
}%
}

\begin{document}

\maketitle
\thispagestyle{empty}
\pagestyle{empty}

\begin{abstract}
This paper adapts a general dataset representation technique to produce robust \Gls{vpr} descriptors, crucial to enable real-world mobile robot localisation.
Two parallel lines of work on \gls{vpr} have shown, on one side, that general-purpose off-the-shelf feature representations can provide robustness to domain shifts, and, on the other, that fused information from sequences of images improves performance.
In our recent work on measuring domain gaps between image datasets, we proposed a \gls{vdna} representation to represent datasets of images.
This representation can naturally handle image sequences and provides a general and granular feature representation derived from a general-purpose model. 
Moreover, our representation is based on tracking neuron activation values over the list of images to represent and is not limited to a particular neural network layer, therefore having access to high- \textit{and} low-level concepts.
This work shows how \glspl{vdna} can be used for \gls{vpr} by learning a very lightweight and simple encoder to generate task-specific descriptors.
Our experiments show that our representation can allow for better robustness than current solutions to serious domain shifts away from the training data distribution, such as to indoor environments and aerial imagery.
\end{abstract}

\glsresetall

\section{Introduction}

\Gls{vpr} is an important task in robotics~\cite{lowry_visual_2016}.
It consists of recognising whether a place has already been observed from an image depicting it.
Doing so robustly and efficiently can help find loop closures for \gls{slam} or directly localise a robot.
However, for a \gls{vpr} system to be useful, it must be robust to viewpoint and appearance changes between different observations of the same place.

This robustness has often been achieved by training systems specifically for \gls{vpr}; however, recently, more robust and general place recognition has been achieved by exploiting general-purpose feature representations~\cite{keetha2023anyloc}.
Another parallel avenue for improving \gls{vpr} performance is to leverage an intrinsic characteristic of the robotics settings: robots continuously stream data, allowing the application of sequences of images to perform place recognition \cite{mereu2022seqvlad}.

On the other hand, a recent research area is studying general tools to measure custom domain gaps between image datasets~\cite{ramtoula2023vdna}.
Here, the idea is to generate a general granular representation of an image dataset and to compare these representations through task-dependent distances.
The two aforementioned  key insights to improve \gls{vpr} are naturally handled in these approaches: representations are selected to be general and robust to domain variations, and representations are generated for an arbitrary number of images.

\begin{figure}
\centering
\includegraphics[width=0.45\textwidth]{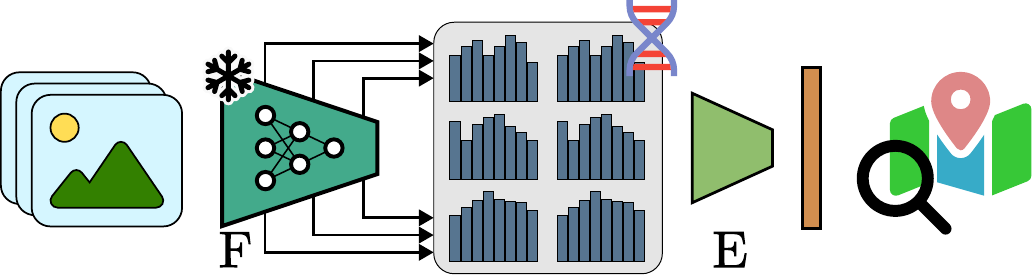}
\caption{
\textit{VDNA-PR overview}.
As in other sequence-based works, we solve \gls{vpr} by building and matching representations for image sequences along driven trajectories. We rely on \acrshort{vdna} representations~\includegraphics[scale=0.09]{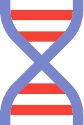}~\cite{ramtoula2023vdna}, which were originally introduced to measure domain gaps between datasets. They consist of histograms that describe activations observed when passing images through a frozen self-supervised feature extractor F.
Importantly, \acrshortpl{vdna} keep track of activations for neurons throughout all layers of the network, keeping a general and granular multi-level representation. To generate more practical descriptors specifically for \gls{vpr}, we propose an encoder E to encode \acrshortpl{vdna} into descriptors that can efficiently be compared with traditional \gls{vpr} techniques.
}
\label{fig:overview}
\vspace{-24pt}
\end{figure}

Thus, this work proposes building a \gls{vpr} system using methodologies borrowed from dataset domain-gap measurements.
A high-level overview is depicted in \cref{fig:overview}.
Specifically, we treat sequences of consecutive images as individual datasets and generate general representations that allow for general-purpose comparisons.
For \gls{vpr}, however, we need to produce practical descriptors that can be compared at scale.
Hence, we learn a ``\gls{vpr} encoder'' on top of the general-purpose representation that produces a small descriptor suitable for the task.
Framing the \gls{vpr} problem in this fashion ensures we have access to a general, robust, and granular representation on top of which to perform \gls{vpr}, and do not have any arbitrary assumption on the choice of the feature extractor layer.

Our principal contributions are
\begin{enumerate}
\item A novel application of deep neural dataset-to-dataset comparison mechanisms to sequence-based \gls{vpr}, 
\item An architecture for learning a more practical descriptor for \gls{vpr}, and
\item Experiments showing improved generalisation under domain shift versus competitors.
\end{enumerate}

\section{Related Works}

\Gls{vpr} is classically set as an image retrieval problem; given a query image and a reference database, its closeness in geometric space to any image in the database is directly related to a similarity measure in an embedding space \cite{lowry_visual_2016}, extracted either from local or global features within the image.
Similarly, images can be processed singularly or sequentially, i.e. exploiting the inherent temporal correlation to create a stronger representation \cite{mereu2022seqvlad}.
This work is related to both challenges of selecting the right representation and aggregating temporal information; thus, we will frame it in both contexts in the following.

\subsection{Selecting the right representation}

With the advent of \gls{dl} methodologies and extensive \gls{vpr} datasets, we have seen an almost complete shift from handcrafted features to learned ones, with substantial performance improvements.
The common approach is to utilise a feature extractor and fine-tune it while training a feature aggregator on top of the learned features to perform \gls{vpr}.
For instance, NetVLAD \cite{arandjelovic2016netvlad} follows this approach, introducing a trainable variant of the VLAD \cite{jegou2010aggregating} descriptor.
This methodology has been very successful and inspired several improvements \cite{zhang2021vector,yu2019spatial} and applications to different data types \cite{suaftescu2020kidnapped,uy2018pointnetvlad}.

Other methods focus on the extraction and aggregation of information from the images.
Examples are MixVPR \cite{ali2023mixvpr} and R2Former~\cite{zhu2023r2former}.
MixVPR uses flattened features from intermediate layers of a pre-trained backbone and incorporates spatial relationships through Feature-Mixer blocks.
R2Former, instead, uses a transformer architecture to encode the image patches into global and local descriptors and uses the first for global retrieval and the local ones for fast reranking through a correlation operation.

GeM \cite{radenovic2018fine} and consequently Cosplace~\cite{berton2022rethinking} focused on data and training procedures, simplifying the aggregation to a learned pooling operation.
Indeed, the first applied a contrastive approach where positives and negatives are selected through a \gls{sfm} technique, whereas the second sets the \gls{vpr} training as a classification problem on a novel extensive dataset. 

Finally, a very recent work, AnyLoc \cite{keetha2023anyloc}, proposes the use of foundational models, in particular DINOv2 \cite{oquab2023dinov2}, pre-trained in a self-supervised fashion and with no prior knowledge about the task of \gls{vpr}, to extract generic and robust features. These features are then exploited for the specific task of \gls{vpr} using an aggregation method, such as VLAD \cite{arandjelovic2016netvlad} or GeM \cite{radenovic2018fine}.
In this way, AnyLoc showed state-of-the-art generalisability on multiple domains.

In this, we are most similar to this last approach, in that we exploit a self-supervised pre-trained model to create a representation of the data robust to domain changes.

\subsection{Using Sequential information}
Sequence-based methods leverage temporal information in the query and database to discover more robust matches for \gls{vpr}. We refer the reader to Mereu \etal{} for a detailed taxonomy~\cite{mereu2022seqvlad}.
Generally, we can divide such methods into two main families: sequence-matching and sequence-descriptor methods.
The firsts use separate similarities for each query image in a query sequence to match \textit{segments} of high similarity in the database.
Methods such as SeqSLAM \cite{milford2012seqslam} and \cite{ho2007detecting} belong to this family, where the first samples and validates feasible trajectories in the database to assess the most likely one and the second uses a matrix value decomposition to eliminate possible noisy detections.

Sequence-descriptor methods, instead, propose to generate a descriptor for an entire image sequence and directly match segments through it.
TimeSformer \cite{bertasius2021space} explores early-fusion approaches on image-patches descriptors through a transformer architecture.
SeqNet \cite{garg2021seqnet}, instead, has a hybrid approach as it uses a convolutional backbone to extract image descriptors, which are used to calculate a sequence-based descriptor for rough sequence matching and to refine this latter through image-to-image similarity.
Mereu \etal~\cite{mereu2022seqvlad} tackle sequential embedding through an intermediate fusion of single-frame descriptors with a SeqVLAD aggregator, a NetVLAD~\cite{arandjelovic2016netvlad} generalisation to handle multiple images. 

Our proposed approach belongs to this last family of methodologies, as we too combine the information from the whole sequence into a single descriptor, and then use it for the matching phase. We base our approach on a previous general dataset representation and comparison work.

\begin{figure*}[!h]
\centering
\includegraphics[width=0.9\textwidth]{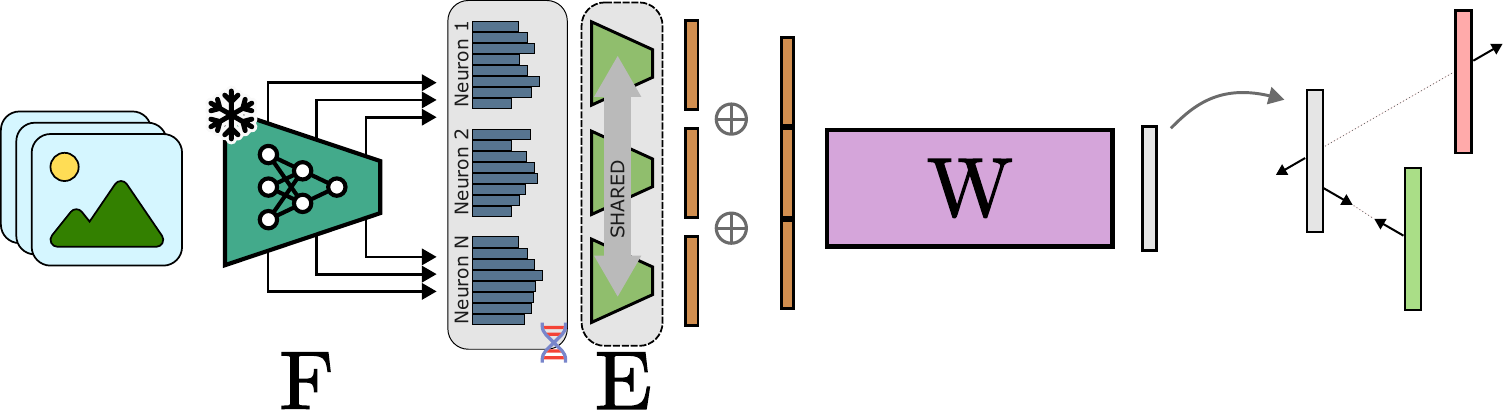}
\caption{
\textit{Overview of VDNA-PR training}.
As in~\cref{fig:overview}, as a sequence of images passes through a pre-trained frozen feature extractor F, histograms tracking neuron-wise activations constitute a \acrshort{vdna}~\includegraphics[scale=0.09]{figs/dna.pdf}.
The histogram corresponding to each neuron has $500$ bins, and a small 1D CNN encoder E maps each histogram to a lower dimensional vector of length $4$ (with shared weights across the $9216$ neurons).
The concatenation of these length-$4$ features is of length $36864$ and is then itself passed through a linear layer W to be reduced in dimension and to form the final representation.
It is on this representation that we perform contrastive learning with triplet losses as is common in place recognition. At test-time on different domains, we remove the linear layer W which has learned specific features of the training domain, and use concatenations of encoded histogram features from selected neurons.
With this training, we therefore learn neuron-wise descriptors that can be used and combined for \gls{vpr}.
}
\label{fig:detailed_method}
\vspace{-14pt}
\end{figure*}

\subsection{Representation scope \& layer combinations}

Our approach benefits from combining representations \textit{across} the feature extractor, and we show experimentally in~\cref{sec:results} that multi-level concepts access in different layers can help generalise better across deployment domains.
Related to this, is the ``filter-early, match-late'' work of~\cite{hausler2019filter} with the guiding principle that simple visual features that detract from a network's utility for place recognition across changing environmental conditions are removed and, as per~\cite{sunderhauf2015performance}, that late layers are more invariant to viewpoint changes.
In our recent work introducing \gls{vdna}~\cite{ramtoula2023vdna} to represent image datasets, we also highlighted the power of granular representations over multiple feature extractor layers to control what concepts the representation comparison should be sensitive to, such as when measuring the realism of synthetic images.
Our combination of layers is also related to a weighted concatenation of convolutional features across layers in~\cite{mao2018learning}.

\subsection{Measuring domain gaps between datasets}
\label{sec:prelims}

Quantifying the domain gap between datasets is possible with techniques such as \gls{fid}~\cite{heusel2017gans}, \gls{kid}~\cite{binkowski2018demystifying}, bag-of-prototypes~\cite{tu2023bag}, and \gls{emd}~\cite{rubner2000earth} applied to \glspl{vdna}~\cite{ramtoula2023vdna}.

\Gls{fid}~\cite{heusel2017gans} embeds images with a specific layer of the Inception-v3 network~\cite{szegedy2016rethinking}, fitting a multivariate Gaussian, and the \gls{fd} between such distributions from different datasets is computed.
Alternatively, the \gls{kid}~\cite{binkowski2018demystifying} computes the squared maximum mean discrepancy between Inception-v3 embeddings.

In addition to relying on a more comprehensive and granular representation, \Gls{vdna}~\cite{ramtoula2023vdna} was demonstrated in~\cite{gadd2023iser} to better predict localisation precision between experiences with serious seasonal domain shifts, particularly in comparison to \gls{fid}~\cite{heusel2017gans}, and so is the representation on which we base our proposed place recognition system.
This method, as illustrated and described in~\cref{fig:overview}, populates neuron-wise histograms with the activation levels of a pre-trained feature extractor as images pass through the network, across all layers.
Histogram comparisons (e.g., by the \gls{emd}) can be aggregated and combined with neuron-wise emphasis (or combinatorially across neurons) to fine-tune the dataset-to-dataset comparison across attributes of interest and at various levels of features in deep networks.

\section{Methodology}

Expanding on~\cref{fig:overview}, our \textit{VDNA-PR} training approach is shown in more detail in~\cref{fig:detailed_method}, with a focus on the lightweight encoder module E learned to produce practical descriptors for the \gls{vpr} problem.

\subsection{Visual DNAs creation and comparisons}
\label{sec:meth:vdna_emd}

We chose to build our representation on top of \gls{vdna}~\cite{ramtoula2023vdna} representations of images from sequences for which we want to produce a \gls{vpr} descriptor.

In this, given a sequence of $L$ images $\mathcal{I}$, where $L \geq 1$, we can pass the images through a frozen feature extractor (\Snowflake~in~\cref{fig:detailed_method}). During the forward pass, we can keep track of the outputs of multiple layers of the model, and decompose each of them into multiple 1D ``neuron activations''. The idea of \gls{vdna} is to form a granular representation by gathering distributions of many neuron activations.    
This is done by collecting neuron activations into histograms $\mathcal{H}_i$, one for each $i$ of the $N$ neurons across all layers.
Each histogram comprises $b$ bins, giving a total descriptor size of $N \times b$ which does not depend on the sequence length $L$ (this being one of the core motivations of the original approach in~\cite{ramtoula2023vdna}).
Indeed, \Glspl{vdna} can inherently represent multiple images simultaneously as they are designed as compact representations of datasets.

As \glspl{vdna} are composed of histograms, the original measure of domain gaps between two datasets represented as \glspl{vdna} is based on the \glsfirst{emd}~\cite{rubner2000earth}.
Unfortunately, while this distance has long been useful in image retrieval~\cite{rubner2000earth}, no efficient implementation is present in commercial-grade image-retrieval database technologies, beyond approximations as in~\cite{atasu2019linear}.

For these reasons, we propose an encoder model, trainable and applied to the task of \gls{vpr}, detailed in the following.
The output of this module is a lower-dimensional vector upon which we can compute $L_2$ distances, at scale.

\subsection{VPR Encoder Model}
\label{sec:meth:encoder}

In typical foundational models, the number of neurons, $N$, may be extremely large.
For instance, DINOv2's \gls{vit} used in this work contains $9216$ neurons\footnote{For comparison, \textit{LeNet5} consists of only $84$ neurons which with $500$ bins per histogram as in~\cref{sec:exp:vdna} already leads to a $42000$-dimensional vector, in excess of VGG-16/NetVLAD~\cite{simonyan2014very,arandjelovic2016netvlad}}.
This fact further supports our need not to directly use the histograms $\mathcal{H}_i$.
Indeed, this descriptor dimension challenges caching it in memory and thus efficient comparison, essential in real-time applications such as robotics.
To tackle these issues and those raised in~\cref{sec:meth:vdna_emd}, we design an encoder model to encode the histograms, normalised, into a latent space where we can efficiently use a vector distance.

This encoder network is shown as the shared modules in~\cref{fig:detailed_method}.
We treat each histogram $\mathcal{H}_i$ as sequential data and apply a 1D \gls{cnn} composed of six convolutional layers followed by three linear layers.
This reduces the dimensionality of each histogram from $b$ to $h$. 
After extensive experimentation, we set $h$ to 4.
After encoding, the extracted embeddings are normalised singularly and concatenated into an embedding $e$ of dimension $Nh$.
Now, this descriptor can more efficiently be stored in memory and compared.

\subsection{Training approach}

Given a labelled dataset of image sequences, we aim to train the encoder model to achieve high-grade accuracy while generalising to different domains.
To achieve this, we take inspiration from  CosPlace training~\cite{berton2022rethinking}, which uses data-specific linear layers on top of the representation of interest that are then discarded after training. We apply a linear layer W, parameterised by a trainable weight matrix $\mathtt{W} \in Nh \times d$, to project the descriptor into a lower dimensionality.
We use these descriptors to produce triplet losses~\cite{hoffer2015deep}, which serve to train the encoder E and linear layer W.
We use a triplet loss as it is popular for \gls{vpr}, but would expect contrastive losses \cite{oord2018representation,chen2020simple} to also work well.

This training will encourage the 1D CNN to produce good encodings of histograms that can be reused on different domains while learning the specificities of the training domain in the linear layer that can be discarded later.

\begin{table}
\centering
\caption{Details of datasets used. MSLS (train) and (val) are used during training, and all others are used for evaluation. The numbers of database and query sequences are given for a sequence length of 5 frames.}
\label{tab:dataset_details}
\begin{tabular}{@{}ccccc@{}}
\toprule
Name  & Domain & \# of db seq. & \# of query seq. & Loc. req. \\ \midrule
MSLS (train)          &      Urban       &       733048        &      393201      &       25 meters               \\
MSLS (val)          &      Urban       &        8125       &     5752       &       25 meters               \\
\midrule
MSLS (test)          &      Urban       &   13584            &  7964          &       25 meters               \\
St Lucia      &      Urban       &       1545        &        1460      &         25 meters          \\
Pitts-30k     &     Urban        &      6664         &        4542      &         25 meters           \\
RobotCar 1    &       Urban      &         3615      &       3292       &           25 meters         \\
RobotCar 2    &      Urban       &         3919      &       3687       &           25 meters         \\
RobotCar 3    &      Urban       &        3628       &      3917        &           25 meters         \\
Baidu Mall    &     Indoor        &      677         &      1356        &        10 meters            \\
Gardens Point &     Indoor        &      196         &     196         &        2 frames            \\
17 Places     &    Indoor         &       334        &     334         &         5 frames           \\
VPAIR        &    Aerial         &       2702        &     2702         &          3 frames          \\ \bottomrule
\end{tabular}
\vspace{-14pt}
\end{table}

\section{Experimental Setup}

\subsection{VDNAs}
\label{sec:exp:vdna}

We use DINOv2 ViT-B/14~\cite{oquab2023dinov2} with $12$ layers of $768$ neurons each to generate \glspl{vdna}.
For each neuron, we collect a histogram of $b=500$ bins, which we find to be sufficient to finely approximate distributions of activations.
We chose this model as AnyLoc~\cite{keetha2023anyloc} showed that DINOv2 achieves great generalisability to different scenes and domains thanks to its training regime.

\subsection{Data}
To validate our approach, we use one training and several testing datasets summarised in \Cref{tab:dataset_details} to evaluate our generalisation capabilities.
In particular, we selected MSLS~\cite{warburg2020msls} to train our neuron encoder module as it contains a large scale of training samples from several urban environments worldwide and with diverse conditions.
Moreover, the data samples are collected as sequences, so they apply to our proposed methodology.

\begin{figure}
\centering
\begin{subfigure}{0.4\columnwidth}\includegraphics[width=\textwidth]{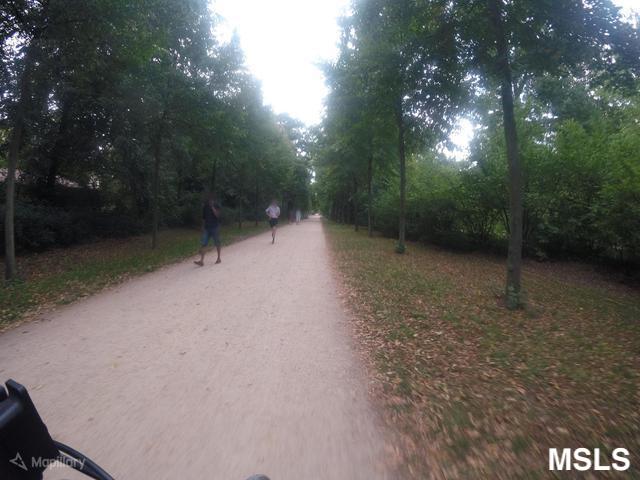}\end{subfigure}
\begin{subfigure}{0.4\columnwidth}\includegraphics[width=\textwidth]{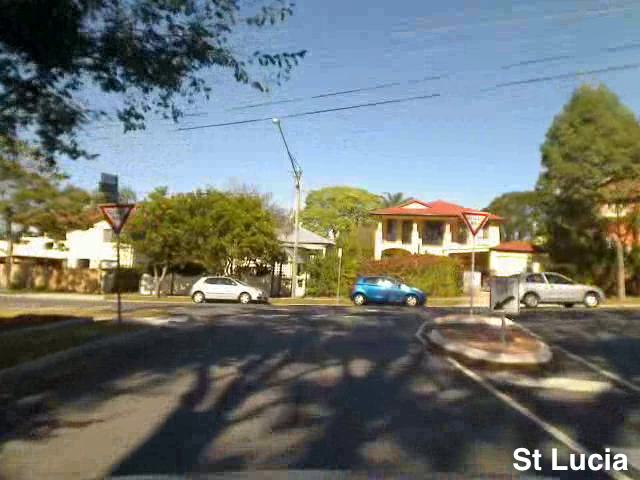}\end{subfigure}\\[5pt]
\begin{subfigure}{0.4\columnwidth}\includegraphics[width=\textwidth]{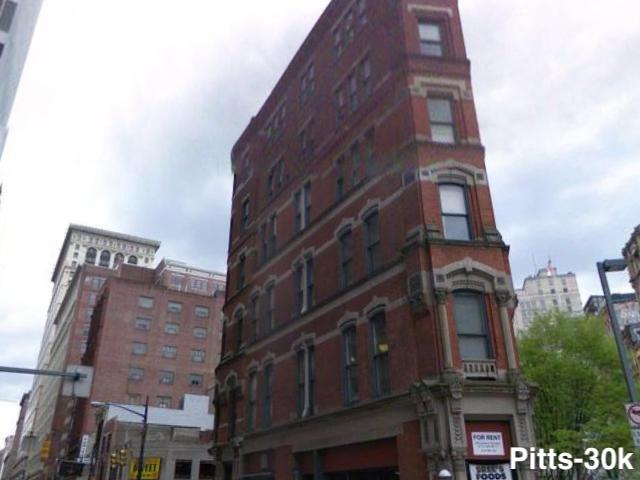}\end{subfigure}
\begin{subfigure}{0.4\columnwidth}\includegraphics[width=\textwidth]{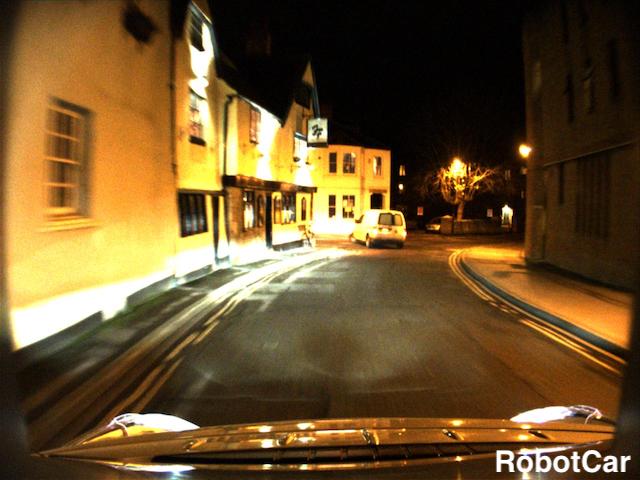}\end{subfigure}\\[5pt]
\begin{subfigure}{0.4\columnwidth}\includegraphics[width=\textwidth]{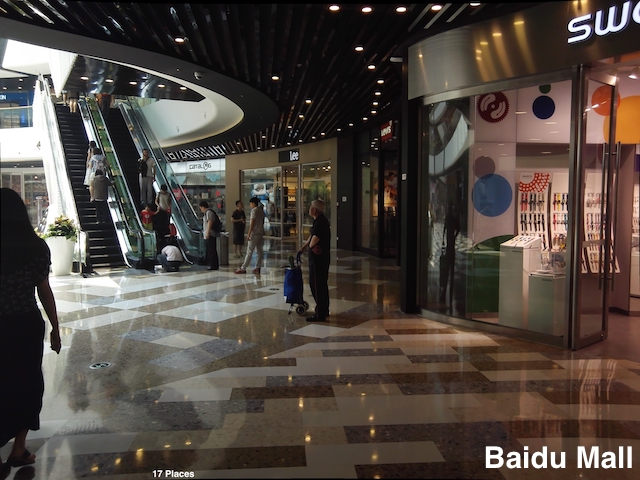}\end{subfigure}
\begin{subfigure}{0.4\columnwidth}\includegraphics[width=\textwidth]{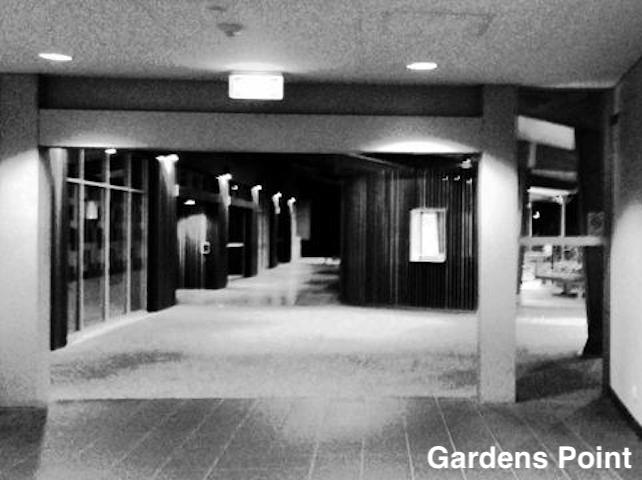}\end{subfigure}\\[5pt]
\begin{subfigure}{0.4\columnwidth}\includegraphics[width=\textwidth]{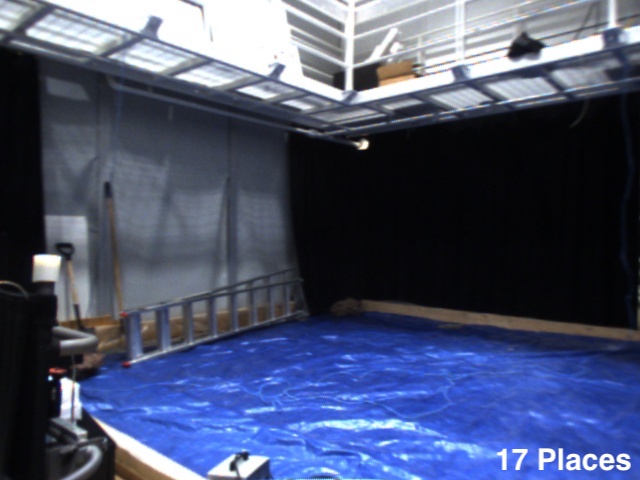}\end{subfigure}
\begin{subfigure}{0.4\columnwidth}\includegraphics[width=\textwidth]{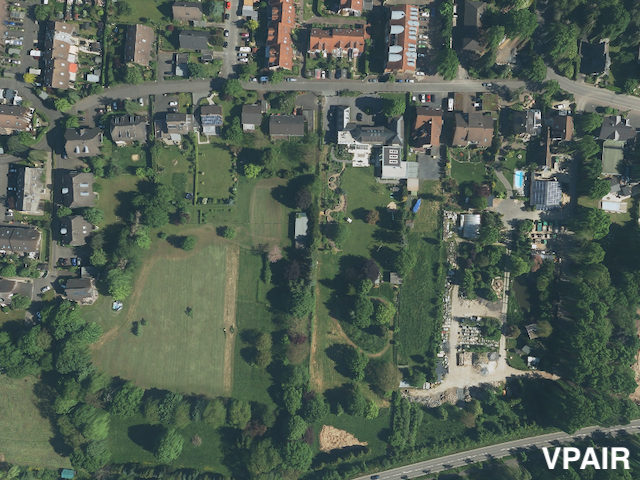}\end{subfigure}
\caption{
Example images from datasets used in this study. We evaluate generalisation capability from training \Gls{vpr} approaches on MSLS, and using them in other urban and indoor environments, as well as on aerial imagery.
}
\label{fig:domains}
\vspace{-14pt}
\end{figure}

We evaluate \gls{vpr} performance on a diverse set of domains, including the \textbf{Baidu Mall Benchmark}~\cite{sun2017dataset}, \textbf{VPAIR}~\cite{schleiss2022vpair}, \textbf{17 Places}~\cite{sahdev2016indoor}, and \textbf{Gardens Point}~\cite{arren_glover_2014_4590133}. 
\Cref{fig:domains} shows examples from the datasets used, which show a high degree of domain variability, from indoor to urban to overhead imagery.
We consider \textbf{Pittsburgh-30k}~\cite{torii2013visual}, \textbf{St Lucia}~\cite{warren2010unaided},  and \textbf{Oxford}~\cite{maddern20171} to have significant domain overlap with \textbf{MSLS}~\cite{warburg2020msls} as they also contain images from urban environments.
In \textbf{Oxford}, we use the three train/val/test splits used in~\cite{mereu2022seqvlad}, corresponding to:
\begin{itemize}
    \item \textbf{RobotCar1}: queries: 2014-12-17-18-18-43 (winter night, rain); database: 2014-12-16-09-14-09 (winter day, sun).
    \item \textbf{RobotCar2}: queries: 2015-02-03-08-45-10 (winter day, snow); database: 2015-11-13-10-28-08 (fall day, overcast).
    \item \textbf{RobotCar3}: queries: 2014-12-16-18-44-24 (winter night); database: 2014-11-18-13-20-12 (fall day).
\end{itemize}

For the training and evaluation datasets, splits and labelling are taken from SeqVLAD \cite{mereu2022seqvlad} and Anyloc \cite{keetha2023anyloc}.
Particularly, we did not consider the distractor images in \textbf{VPAIR} as they are not collected as sequences.

In our experiments, we employ a sequence length of 5 images, which we keep consistent with all our baselines.
Nevertheless, we ablate over this number to investigate how a network pre-trained on five images will behave when used with a different number.

\subsection{Training details}
For training our encoder, we use similar mining methods from previous works~\cite{mereu2022seqvlad,warburg2020msls}. We cache 1000 query sequences alongside 5000 negatives to perform hard example mining. We also include the 500 hardest previous triplets and refresh the cache every 1500 triplets. Each triplet contains a query, its positive, and 5 negatives. We optimise the encoder E and linear layer W weights using the AdamW~\cite{loshchilov2018decoupled} optimiser with weight decay.

\subsection{Baseline}
\label{sec:exp:baselines}
All approaches considered rely on the DINOv2 ViT-B/14~\cite{oquab2023dinov2} backbone. For all \textit{SeqVLAD} variants, we use features from layer 12. Our baselines and variants include:
\begin{enumerate}
\item {\crtcrossreflabel{\texttt{$\text{SeqVLAD}$}}[seqvlad]}: The method and general architecture as described in~\cite{mereu2022seqvlad}, but with DINOv2 substituted as the backbone. 
In this variant, the backbone and the SeqVLAD layer are both fine-tuned on MSLS~\cite{warburg2020msls}, as is done in~\cite{mereu2022seqvlad}.
\item {\crtcrossreflabel{\texttt{$\text{SeqVLAD}_{\text{frozen}}$}}[seqvlad_frozen_encoder]}: The same setup as \ref{seqvlad}, but in this case, only the SeqVLAD aggregation layer is fine-tuned. The backbone is kept frozen.
\item {\crtcrossreflabel{\texttt{$\text{SeqVLAD}_{\text{calib.}}$}}[vlad]}: Using SeqVLAD calibrated to the evaluation domain. This is performed similarly to the use of Hard Assignment VLAD~\cite{arandjelovic2013vlad} in AnyLoc~\cite{keetha2023anyloc} but with the SeqVLAD aggregation over multiple images. The backbone is not fine-tuned, and centroids are initialised using clustering on the database images. 
\item {\crtcrossreflabel{\texttt{$\text{VDNA-PR}_{\mathtt{W}}$}}[vdna_trained_linear_layer]}: Using the output of the linear layer W used during the training of \textit{VDNA-PR}.
\item {\crtcrossreflabel{\texttt{$\text{VDNA-PR}_{\text{all}}$}}[vdna_all_neurons]}: Using the output of the \textit{VDNA-PR} encoder E (i.e. with the W linear layer omitted) when \textit{all} $9216$ neurons have been projected to length $4$ and concatenated. This is equivalent to \ref{vdna_trained_linear_layer} with the linear layer used during training on MSLS removed.
\item {\crtcrossreflabel{\texttt{$\text{VDNA-PR}_{12}$}}[vdna_layer_12]}: Using the \textit{VDNA-PR} encoder E output but only keeping descriptors from neurons in the twelfth layer.
\item {\crtcrossreflabel{\texttt{$\text{VDNA-PR}_{9:12}$}}[vdna_layer_9_to_12]}, {\crtcrossreflabel{\texttt{$\text{VDNA-PR}_{10:12}$}}[vdna_layer_10_to_12]}, and {\crtcrossreflabel{\texttt{$\text{VDNA-PR}_{11:12}$}}[vdna_layer_11_to_12]}: Similar to \ref{vdna_layer_12} but in these cases with a \textit{combination} of layers as subscripted.
\end{enumerate}

Baselines (1-3) produce a descriptor of length 49152, (4) produces a descriptor of length 128, and (5-7) produce descriptors of length $3072\times$\#\texttt{layers}.

\subsection{Performance metrics}

To assess place recognition performance, we use the {Recall@N (R@N)} metric. It consists in the percentage of successful queries in the set of query images which result in at least $1$ correct localisation result when the $N$ closest reference images are retrieved from the reference map or database according to \gls{vpr} descriptors.
``Correctness'' of a localisation match is determined by a ground-truth distance -- these thresholds are detailed in~\cref{tab:dataset_details}.
\section{RESULTS}
\label{sec:results}
\subsection{Effect of layer selection}
\begin{figure}
\centering
\includegraphics[width=0.4\textwidth]{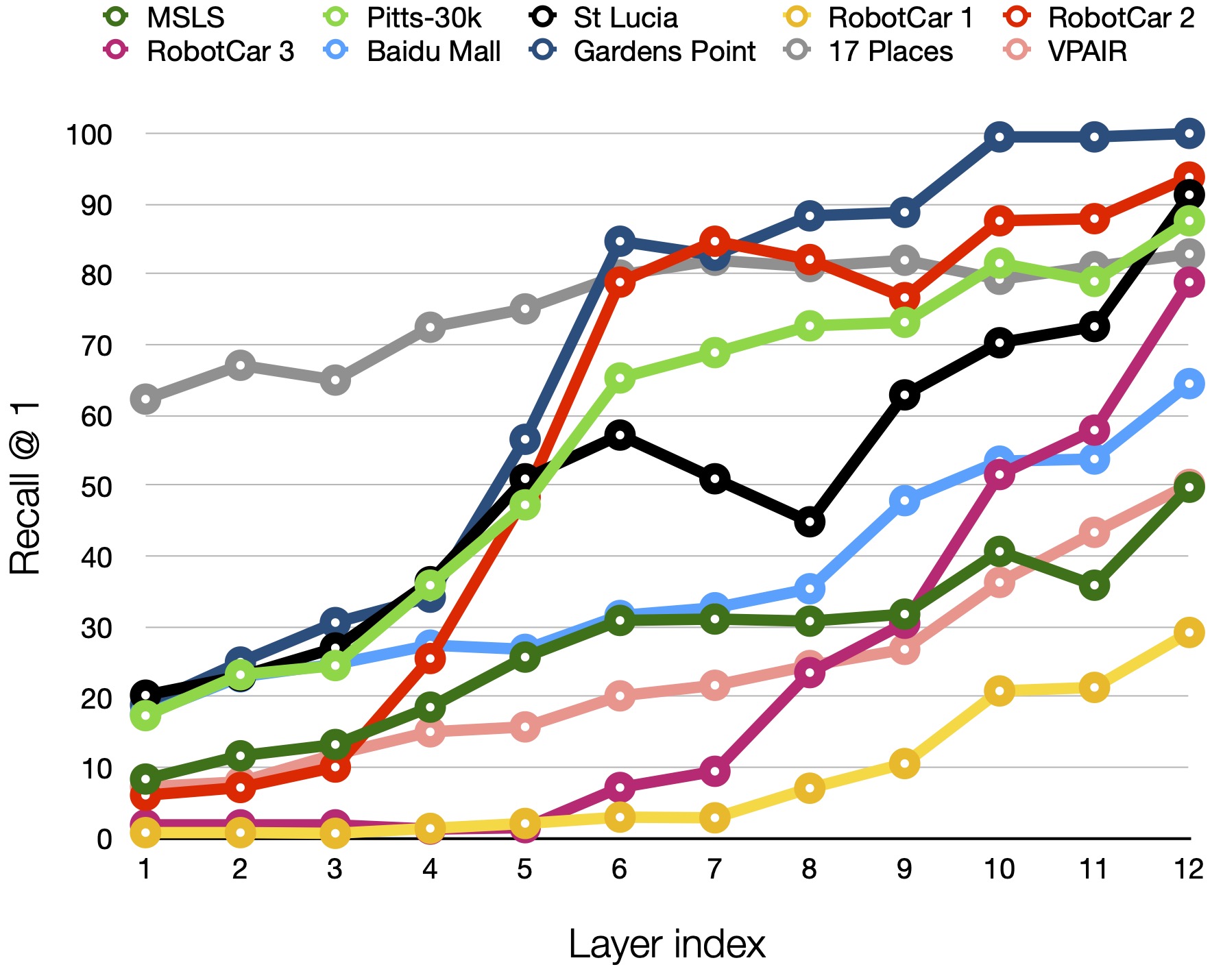}
\caption{Recall@1 when evaluating \gls{vpr} performance using \textit{VDNA-PR} descriptors from each layer of DINOv2 for all datasets considered.
}
\label{fig:layers_vs_recalls}
\vspace{-10pt}
\end{figure}

In \cref{fig:layers_vs_recalls}, we first observe how using a \textit{VDNA-PR} descriptor composed of histogram encodings of neurons from each backbone layer affects performance on all datasets considered. 

We obtain large variations depending on the datasets. For example, decent performance on \textbf{17 Places} can be obtained even with early layers, but not on \textbf{RobotCar} sequences. Moreover, a later layer does not always lead to improved performance. For example, we obtained better results on \textbf{St Lucia} with neurons from layer 6 than from layer 8. These observations support the value of having a descriptor containing information from all layers, as not all datasets will do best with the same levels of features. Overall, the last layers consistently lead to the best results for all datasets. Hence, in our following experiments, we include results based on neurons from combinations of the last layers.

\subsection{Benchmarking with domain shifts}

\begin{table*}[!h]
	\centering
	\caption{Performance comparison on benchmark environments with urban and indoor environments, and aerial imagery.
		Please refer to the baseline/ablation definitions in~\cref{sec:exp:baselines} for details of methods (row names).
	}
	\resizebox{\textwidth}{!}{
		\begin{tabular}{ccccccccccccccccccccc}
			\toprule
			& \multicolumn{12}{c}{\textbf{Urban}} & \multicolumn{6}{c}{\textbf{Indoor}}  & \multicolumn{2}{c}{\textbf{Aerial}}\\
                \cmidrule(lr){2-13} \cmidrule(lr){14-19} \cmidrule(lr){20-21} 
			& \multicolumn{2}{c}{\textbf{MSLS}} & \multicolumn{2}{c}{\textbf{St Lucia}} & \multicolumn{2}{c}{\textbf{Pitts-30k}} & \multicolumn{2}{c}{\textbf{RobotCar 1}} & \multicolumn{2}{c}{\textbf{RobotCar 2}} & \multicolumn{2}{c}{\textbf{RobotCar 3}} & \multicolumn{2}{c}{\textbf{Baidu Mall}} & \multicolumn{2}{c}{\textbf{Gardens Point}} & \multicolumn{2}{c}{\textbf{17 Places}} & \multicolumn{2}{c}{\textbf{VPAIR}} \\
			  \textbf{Methods}                              & R@1           & R@5           & R@1            & R@5            & R@1           & R@5           & R@1           & R@5           & R@1           & R@5           & R@1           & R@5           & R@1           & R@5           & R@1           & R@5           & R@1           & R@5           & R@1           & R@5           \\
		\cmidrule(lr){1-1} 	\cmidrule(lr){2-3} \cmidrule(lr){4-5} \cmidrule(lr){6-7} \cmidrule(lr){8-9} \cmidrule(lr){2-3} \cmidrule(lr){4-5} \cmidrule(lr){6-7} \cmidrule(lr){8-9} \cmidrule(lr){10-11}  \cmidrule(lr){12-13}  \cmidrule(lr){14-15}  \cmidrule(lr){16-17}  \cmidrule(lr){18-19}  \cmidrule(lr){20-21}
			\ref{seqvlad}                    & \textbf{87.9} & \textbf{91.7} & \textbf{99.7} & \textbf{99.9} & 92.8          & 96.1          & 35.5          & 45.1          & 97.0          & 98.1          & 68.3          & 76.9          & 52.1          & 69.5          & 89.8           & 95.9           & 81.1          & 86.8          & 20.5          & 30.5          \\
			\ref{seqvlad_frozen_encoder}     & 82.2          & 87.6          & 99.5          & \textbf{99.9} & \textbf{93.4} & \textbf{96.6} & \textbf{66.0} & \textbf{74.1} & \textbf{98.3} & \textbf{98.9} & \textbf{93.3} & \textbf{96.3} & 63.5          & 80.4          & \textbf{100.0} & \textbf{100.0} & \textbf{83.5} & \textbf{89.8} & 44.4          & 57.2          \\
			\ref{vlad}                       & 44.7          & 57.2          & 62.6          & 76.8          & 76.7          & 89.3          & 33.0          & 51.4          & 86.1          & 95.1          & 58.0          & 77.1          & 50.9          & 73.4          & 97.4           & \textbf{100.0} & 81.7          & 87.7          & 20.7          & 34.5          \\
			\ref{vdna_trained_linear_layer}  & 74.1          & 84.0          & 92.1          & 95.3          & 81.2          & 93.2          & 34.7          & 49.9          & 92.2          & 96.8          & 69.8          & 86.2          & 37.1          & 63.8          & 88.3           & 95.4           & 78.7          & 88.6          & 24.8          & 41.7          \\
			\ref{vdna_all_neurons}           & 38.9          & 46.0          & 79.2          & 86.1          & 81.7          & 88.7          & 8.6           & 19.6          & 91.0          & 94.5          & 51.9          & 72.4          & 49.7          & 72.8          & 99.0           & \textbf{100.0} & 80.2          & \textbf{89.8} & 40.4          & 50.4          \\
			\ref{vdna_layer_12}              & 49.8          & 58.5          & 91.3          & 95.1          & 87.6          & 94.4          & 29.2          & 40.6          & 93.8          & 96.5          & 78.9          & 88.1          & 64.5          & \textbf{81.5} & \textbf{100.0} & \textbf{100.0} & 82.9          & \textbf{89.8} & 50.2          & 63.5          \\
			\ref{vdna_layer_9_to_12}         & 45.4          & 52.6          & 86.1          & 91.7          & 85.4          & 92.0          & 30.6          & 42.5          & 94.6          & 97.3          & 78.5          & 88.4          & 66.0          & 81.2          & \textbf{100.0} & \textbf{100.0} & 82.3          & 89.2          & \textbf{51.3} & 63.2          \\
			\ref{vdna_layer_10_to_12}        & 46.0          & 54.1          & 87.1          & 92.2          & 86.0          & 92.6          & 29.9          & 41.4          & 94.9          & 97.4          & 77.9          & 87.8          & \textbf{66.2} & 81.3          & 99.5           & \textbf{100.0} & 82.0          & 89.2          & 51.1          & 63.7          \\
			\ref{vdna_layer_11_to_12}        & 46.7          & 54.5          & 88.6          & 93.6          & 86.2          & 92.9          & 28.3          & 40.9          & 93.7          & 96.7          & 78.3          & 87.9          & 65.4          & 81.4          & \textbf{100.0} & \textbf{100.0} & 82.9          & 89.5          & 51.0          & \textbf{64.3} \\
		\bottomrule
		\end{tabular}
	}
	\label{tab:combined_urban_nonurban_results}
 \vspace{-7pt}
\end{table*}	
\begin{table}[!h]
	\centering
	\caption{Performance comparison on \textbf{VPAIR} with varying imbalanced test-time sequence lengths. All approaches were trained on sequences of length 5 for database and query sequences. Here we denote test-time sequence lengths as database/queries.}
	\resizebox{\columnwidth}{!}{
		\begin{tabular}{ccccccc}
            \toprule
            & \multicolumn{2}{c}{\textbf{1/5}} & \multicolumn{2}{c}{\textbf{5/1}} & \multicolumn{2}{c}{\textbf{5/5}} \\
			       \textbf{Methods}                      & R@1           & R@5           & R@1           & R@5           & R@1           & R@5             \\
			\cmidrule(lr){1-1} \cmidrule(lr){2-3} \cmidrule(lr){4-5} \cmidrule(lr){6-7} 
			\ref{seqvlad}                & 17.31         & 33.5          & 15.4          & 24.7          & 20.5          & 30.5         \\
			\ref{seqvlad_frozen_encoder} & 38.8          & 59.0          & 39.0          & 51.7          & 44.4          & 57.2            \\
			\ref{vlad}                   & 10.3          & 24.9          & 8.8           & 20.3          & 20.7          & 34.5               \\
			\ref{vdna_layer_12}          & \textbf{45.4} & \textbf{66.8} & \textbf{44.7} & \textbf{58.0} & \textbf{50.2} & \textbf{63.5}  \\
            \bottomrule
		\end{tabular}
	}
	\label{tab:vpair_results_1v5}
\end{table}

\begin{table}[!h]
	\centering
	\caption{Performance comparison on \textbf{VPAIR} with increasing test-time sequence lengths. All approaches were trained on sequences of length 5 for database and query sequences.}
	\resizebox{\columnwidth}{!}{
		\begin{tabular}{ccccccc}
                \toprule
                &  \multicolumn{2}{c}{\textbf{seq. len. 5}} & \multicolumn{2}{c}{\textbf{seq. len. 15}} & \multicolumn{2}{c}{\textbf{seq. len. 25}}\\
			       \textbf{Methods}                           & R@1           & R@5           & R@1           & R@5           & R@1           & R@5           \\
			\cmidrule(lr){1-1} \cmidrule(lr){2-3} \cmidrule(lr){4-5} \cmidrule(lr){6-7} 
			\ref{seqvlad}                     & 20.5          & 30.5          & 27.9          & 35.9          & 32.2          & 39.0          \\
			\ref{seqvlad_frozen_encoder}        & 44.4          & 57.2          & 55.4          & 61.8          & 61.7          & 65.8          \\
			\ref{vlad}                         & 20.7          & 34.5          & 27.1          & 35.8          & 34.8          & 41.2          \\
			\ref{vdna_layer_12}         & \textbf{50.2} & \textbf{63.5} & \textbf{61.2} & \textbf{66.7} & \textbf{64.5} & \textbf{68.0} \\
            \bottomrule
		\end{tabular}
	}
	\label{tab:vpair_results_seq_len}
 \vspace{-14pt}
\end{table}

\paragraph{Training domain}

\cref{tab:combined_urban_nonurban_results} (\textit{Urban}) shows the R@1 and R@5 performance of our baselines and approaches in the same domain as they were trained, in urban environments. On the test set of MSLS, \ref{seqvlad} with its fine-tuned backbone performs best. \ref{seqvlad_frozen_encoder} and \ref{vdna_trained_linear_layer} also perform well, thanks to their training on MSLS and despite keeping their backbones frozen.
Other \textit{VDNA-PR} variants without the linear layer and \ref{vlad} do not benefit from a strong specialisation on MSLS, and these approaches perform reasonably well, but noticeably worse than the ones specialised on MSLS.

However, in other urban datasets, \ref{seqvlad} does not dominate as strongly. It particularly struggles on \textbf{RobotCar1}, performing similarly to \textit{VDNA-PR} techniques without W and \ref{vlad}, whereas \ref{seqvlad_frozen_encoder} maintains good performance on other urban domain datasets.
This already indicates a limitation in robustness by using an approach too strongly trained on a dataset, despite MSLS' large scale and diversity.
The features from the selected layer of the backbone might already be well-suited for \gls{vpr} in urban settings. In this case, keeping the backbone frozen allows \ref{seqvlad_frozen_encoder} to outperform \ref{seqvlad} on other urban datasets by maintaining these features.
If the features from a layer are already suitable for \gls{vpr} within a domain, we also expect \ref{seqvlad_frozen_encoder} to be more descriptive than \textit{VDNA-PR} variants as \glspl{vdna} do not have information on the joint distribution of all neuron activations within a layer.

\paragraph{Shifted domains}
\cref{tab:combined_urban_nonurban_results} (\textit{Indoor \& Aerial}) shows the R@1 and R@5 performance of our system on domains different from training: indoors with \textbf{Baidu Mall}, \textbf{Gardens Point} and \textbf{17 Places}, and on aerial imagery with \textbf{VPAIR}.
What we see is that our \textit{VDNA-PR} variants are typically more resilient than other methods, for these datasets.
In \textbf{Gardens Point} and \textbf{17 Places}, we are broadly in line with \ref{seqvlad_frozen_encoder} but outstrip it for \textbf{Baidu Mall} and \textbf{VPAIR}. 

Our \ref{vdna_trained_linear_layer} variant struggles in these non-urban domains.
This is to be expected as we have fine-tuned the linear layer W on \textbf{MSLS} data in \ref{vdna_trained_linear_layer}, and as a result of this urban specialisation, \ref{vdna_trained_linear_layer} suffers in non-urban environments.
On the other hand, \ref{vdna_all_neurons}, which simply removes the linear layer from \ref{vdna_trained_linear_layer}, performs worse in urban domains, but is much more robust to other domains. In fact, the \textit{VDNA-PR} variants lacking the linear layer W are \textit{the most} robust in these datasets, outperforming \textit{SeqVLAD} baselines.
Therefore, we can argue that the fine-tuning of \ref{seqvlad} and \ref{seqvlad_frozen_encoder} on MSLS, despite the foundation model's self-supervised representations, leads to domain shift susceptibility, and in this context our \gls{vdna} representation and careful training and then dismantling of our \gls{vpr} system immunises us from this to a good extent.

We also note that \ref{vlad} always performs worse than \ref{seqvlad_frozen_encoder}, suggesting that the aggregation layer optimisation with a frozen backbone still allowed for some robustness to domain shifts.

Finally, in the case of \textbf{Baidu Mall} and \textbf{VPAIR}, these results show that incorporating information across neural network layers as is natural for \textit{VDNA-PR}  is important, where we see that \ref{vdna_layer_10_to_12} is the most performant in \textbf{Baidu Mall} for R@1 while  \ref{vdna_layer_9_to_12} and \ref{vdna_layer_11_to_12} are the most performant in R@1/5 respectively for \textbf{VPAIR}.
Indeed, considering R@5 for \textbf{17 Places} and \textbf{Gardens Point}, we see that incorporating all neurons is beneficial.
    
Layer combinations considered in this work are arbitrarily chosen and are unlikely to focus on the best set of neurons for each domain. However, these first results are encouraging signals that future work identifying how to select relevant neurons for a given domain could make \textit{VDNA-PR} an even stronger solution for robust \gls{vpr}.

\subsection{Test-time sequence length variations}
Furthermore, we investigate the effect of having varying numbers of images for the database and query descriptors, in particular for the \textbf{VPAIR} dataset.
For \textit{VDNA-PR}, the inputs to the encoder E are normalised histograms regardless of the number of images. This allows to make descriptors consistent for arbitrary numbers of images.

As with SeqVLAD, we can verify this natural robustness when having 1 or 5 sequence lengths in \cref{tab:vpair_results_1v5}. The small variations in performance observed when using 1 and 5 database or query images can be attributed to less informative descriptors due to the reduced number of images used.

In \cref{tab:vpair_results_seq_len}, we also observe that increasing the number of frames is handled naturally and allows for improved performance thanks to more informative descriptors.

\section{CONCLUSIONS}

We have presented a new approach to \gls{vpr} based on a general dataset representation called \acrlong{vdna}.
This representation lends itself naturally to representing sequences and is convenient for combining representations at different layers.
We showed improved resilience of localisation performance to serious domain shifts in non-urban scenes, making progress in the area of universal place recognition from a foundation model basis.

In the future, we will explore the potential of this method with unsupervised \textit{domain calibration} -- where a search for responsive neurons may replace difficult gradient-based fine-tuning of aggregation layers. This could be used on database images to calibrate attention on neurons from which to produce a \gls{vpr} descriptor.






\bibliographystyle{IEEEtran}
\bibliography{bib}

\end{document}